\documentclass[reprint,showpacs,onecolumn,superscriptaddress,aps,pra]{revtex4-2}

\usepackage[utf8]{inputenc}
\usepackage[american,british]{babel}
\usepackage[T1]{fontenc}
\usepackage[pdftex]{graphicx}  
\usepackage{graphicx, xcolor}
\usepackage{dcolumn}
\usepackage{physics}
\usepackage{bm}
\usepackage{amsmath,amsthm,amssymb}
\usepackage{color}
\usepackage{verbatim} 
\usepackage{soul}

\usepackage{multirow}

\usepackage[normalem]{ulem}
\usepackage{natbib}

\newcommand{\smt}{Science, Mathematics and Technology Cluster, Singapore University of Technology and Design, 8 Somapah Road, 487372 Singapore}
\newcommand{\epd}{Engineering Product Development Pillar, Singapore University of Technology and Design, 8 Somapah Road, 487372 Singapore}
\newcommand{\cqt}{Centre for Quantum Technologies, National University of Singapore 117543, Singapore}

\newcommand{\htx}{Home Team Science and Technology Agency, 1 Stars Ave, 138507 Singapore}
\newcommand{\ste}{Artificial Intelligence and Data Analytics Strategic Technology Centre, ST Engineering}
    
\definecolor{darkGreen}{RGB}{0,110,0}
\definecolor{darkBlue}{RGB}{0,0,130}
\usepackage[hidelinks]{hyperref}

\hypersetup{
 colorlinks=true,
 linkcolor=blue,
 citecolor = magenta,
 urlcolor = blue,
pdfborder={0 0 0},
}

\usepackage{dsfont}

\begin{document}
\preprint{APS/123-QED}
\title{SMT-AD: a scalable quantum-inspired anomaly detection approach}

\author{Apimuk Sornsaeng}
\affiliation{\smt}
\affiliation{\cqt}

\author{Si Min Chan}
\affiliation{\cqt} 
\affiliation{\ste}

\author{Wenxuan Zhang} 
\affiliation{\smt}
\affiliation{\cqt} 

\author{Swee Liang Wong}
\affiliation{\htx} 

\author{Joshua Lim}
\affiliation{\htx} 

\author{Jonathan Pan}
\affiliation{\htx}  

\author{Dario Poletti}
\email{dario\_poletti@sutd.edu.sg}
\affiliation{\smt}
\affiliation{\cqt}
\affiliation{\epd}

\date{\today}
\begin{abstract} 
Quantum-inspired tensor network algorithms have shown to be effective and efficient models for machine learning tasks, including anomaly detection. Here, we propose a highly parallelizable quantum-inspired approach which we call SMT-AD from Superposition of Multiresolution Tensors for Anomaly Detection. It is based on the superposition of bond-dimension-1 matrix product operators to transform the input data with Fourier-assisted feature embedding, where the number of learnable parameters grows linearly with feature size, embedding resolution, and the number of additional components in the matrix product operators structure. We demonstrate successful anomaly detection when applied to standard datasets, including credit card transactions, and find that, even with minimal configurations, it achieves competitive performance against established anomaly detection baselines. Furthermore, it provides a straightforward way to reduce the weight of the model and even improve the performance by highlighting the most relevant input features.
\end{abstract}

\maketitle

\section{Introduction}

Anomaly detection is a fundamental problem in machine learning, with applications ranging from fraud detection and cybersecurity to healthcare and industrial monitoring \cite{chandola2009anomaly,pang2021deep}. The goal is to identify rare or atypical samples that deviate from the dominant population of normal data. In many practical scenarios, anomalous samples are scarce, heterogeneous, and often unavailable during training, leading naturally to a one-class learning setting, which is the focus of this work, in which models are trained only on normal data and must detect anomalies as deviations from the learned notion of normality.    

A wide range of approaches has been developed for this task. One-class support vector machines (OC-SVM) aim to learn a boundary enclosing normal data \cite{scholkopf2001estimating}, while isolation-based methods such as Isolation Forest (IF) detect anomalies based on their susceptibility to random partitioning \cite{liu2008isolation,liu2012isolation}. Deep learning approaches have achieved strong empirical performance by learning representations of normal data, for instance using autoencoders \cite{xu2015appearance, andrews2016autoencoder, seebock2019unsupervised}, deep belief networks \cite{erfani2016high}, generative adversarial networks \cite{goodfellow2014gan, schlegl2017unsupervised, zenati2018efficient, donahue2016adversarial, akcay2019ganomaly, deecke2018gan} and transformations, like in GOAD \cite{bergman2020classification}. 

Tensor networks provide a promising framework for addressing this problem. Originally developed in quantum many-body physics, tensor networks such as matrix product states (MPS) offer compact representations of high-dimensional objects with controlled complexity \cite{white1992dmrg, schollwock2005dmrg, schollwock2011dmrg, orus2014tensor}. Their application to machine learning has demonstrated that they can efficiently encode nonlinear feature maps and capture structured correlations with favorable scaling properties \cite{stoudenmire2016supervised,efthymiou2019tensornetwork, HanZhang2018, Guo2018MPOlearning, Guo2020NonMarkovian, Casagrande2024PCA, novikov2017exponentialmachines, oseledets2011tt, cichocki2014erabigdataprocessing}. These properties make tensor networks particularly attractive for anomaly detection, where one seeks to model the structure of normal data while maintaining computational efficiency and interpretability as shown in \cite{wang2020anomaly, aizpurua2025tensor, ZUNKOVIC2023126556}. In particular, the tensor-network anomaly detection (TNAD) framework \cite{wang2020anomaly} demonstrated that matrix product operator (MPO) models can learn one-class decision functions from normal data alone while remaining competitive with standard baselines. However, existing approaches often rely on sequential optimization procedures, which can limit scalability and parallelization.

In this work, we introduce SMT-AD, from Superposition of Multiresolution Tensors for Anomaly Detection. SMT-AD combines three key ideas: a rank-based preprocessing that robustly normalizes individual features; a Fourier-assisted multiresolution embedding that maps each input into a product-state MPS; and a lightweight model built as a superposition of bond-dimension-one MPOs. The model is trained only on normal data, and assigns each input a normality score defined by the overlap of the resulting output state with a fixed reference product state. In this way, normal samples are mapped close to the reference state, while anomalous inputs are detected as deviations from the learned normal manifold.
The proposed construction leads to a highly compact parametrization. In particular, the number of learnable parameters grows linearly with the number of features, the number of embedding resolutions via Fourier modes, and the number of superposed MPO components. This yields a model that is highly parallelizable and vectorizable, making it attractive for low-end hardware, edge computing, and other efficiency-critical environments. At the same time, the superposition structure and multiresolution embedding provide sufficient expressive power for effective anomaly detection.
We benchmark SMT-AD on five standard tabular datasets: \textit{Wine}, \textit{Lymphography}, \textit{Thyroid}, \textit{Satellite}, and \textit{Credit Card}. Across these benchmarks, SMT-AD achieves consistently strong performance, matching or exceeding OC-SVM, IF, and TNAD in the area under the receiver operating characteristic curve (AUROC) on all datasets, while remaining competitive in the area under the precision-recall curve (AUPRC). We also show that the embedding resolution acts as a calibration mechanism for the normality score, with intermediate Fourier modes providing the clearest separation between normal and anomalous samples. An additional strength of SMT-AD is its interpretability. Because the model has an explicit tensor-network structure, one can analyze the learned representation using quantum-information-inspired quantities. In particular, we show that local entropy of the states can be used to identify features that are most relevant for distinguishing anomalous from normal samples, and we use this to improve the performance of anomaly detection while even reducing the size of the model. 

The paper is organised as follows. In Sec. \ref{Section: Model}, we describe the model that we designed, including preprocessing steps, embedding of features to MPS, and the classification MPO. In Sec. \ref{Section: Experiments}, we report on our implementation and results, providing an analysis of the improved performance of SMT-AD compared to other anomaly detection models. We then analyze \textit{how} the model works in Sec.~\ref{section: analysis}, where we consider the feature importance, feature-feature correlation, and resource complexity of the model. 
Finally, we summarize the findings in Sec.~\ref{section: conclusion}. 

\section{Model}\label{Section: Model}

\begin{figure}
    \centering
    \includegraphics[width=1.0\linewidth]{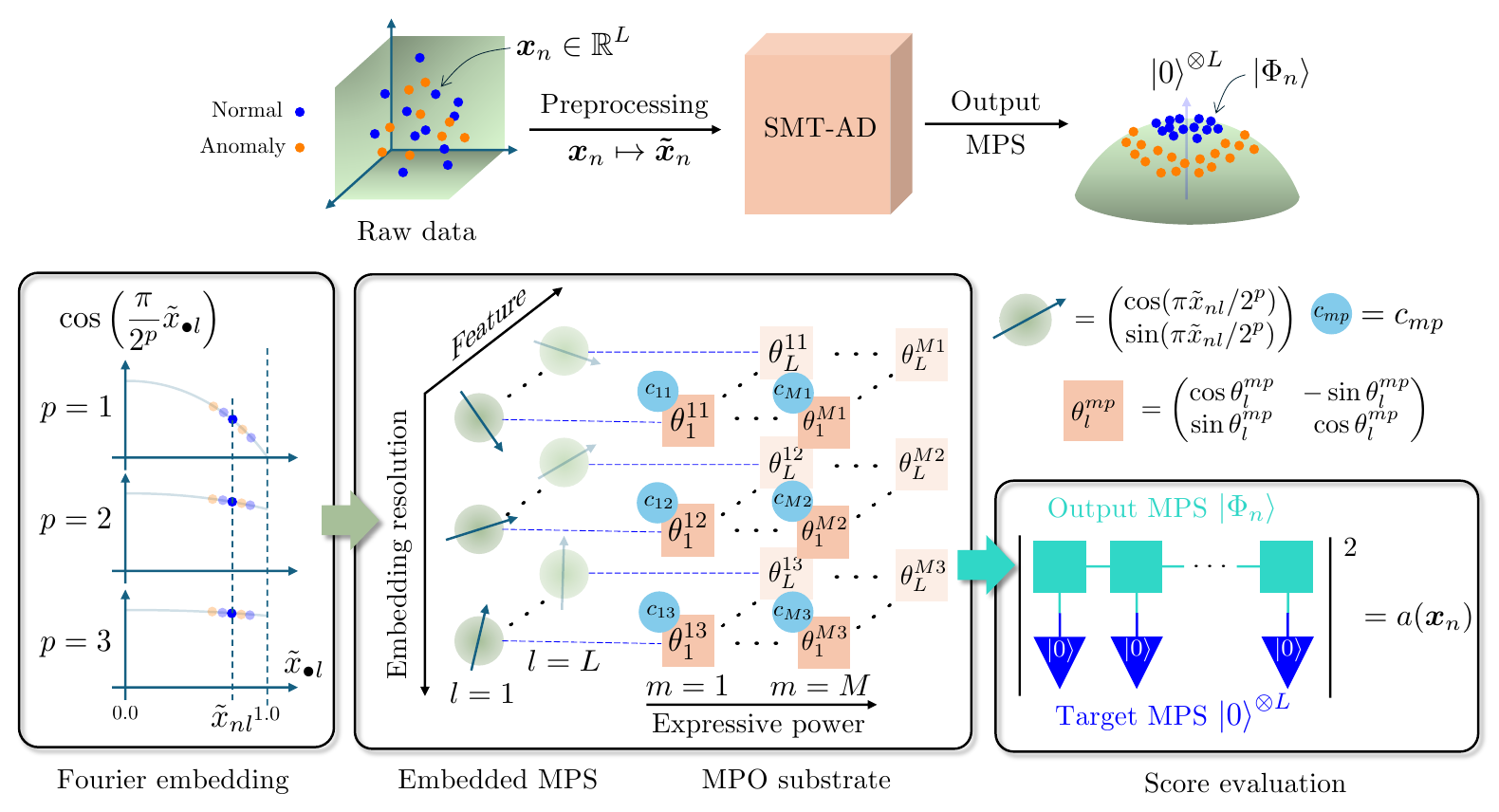}
    \caption{Schematic workflow of the anomaly detection with SMT-AD. The $L$-dimensional input vectors $\{\vb*{x}_n\}_{n=1}^N$ are scaled to $[0,1]$ in the preprocessing step. Nonlinearity is then introduced for each feature $l$ (valued by $\tilde{x}_{\bullet l}$) via Fourier embedding across $P$ frequencies (illustrated here with $P=3$), mapping the data into an input MPS. The trained MPO substrate---comprising a superposition of $MP$ rank-1 MPOs---transforms the input MPS to distinguish anomalous from normal samples. Classification is based on a normality score $a(\vb*{x}_n)$, calculated as the squared overlap of the resulting output MPS $\ket{\Phi_n}$ and a reference target MPS $\ket{0}^{\otimes L}$.}
    \label{fig: SMT_AD}
\end{figure}

Let $\mathcal{D} = \qty{(\vb*{x}_n, y_n)}_{n=1}^N$ denote a dataset for binary classification, where $\vb*{x}_n = \qty(x_{n1},\ldots,x_{nL})\in\mathbb{R}^{L}$ is a raw input having $L$ features, and $y_n \in \{0,1\}$ is its associated label. In the context of anomaly detection, we interpret $y_n=0$ as normal (negative) data and $y_n=1$ as anomalous (positive) data, and accordingly decompose the dataset into $\mathcal{N}\subset \mathcal{D}$ and $\mathcal{A}= \mathcal{D}\setminus\mathcal{N}$, respectively. The main concept of SMT-AD is that a reliable tensor-network-based model for classification can be learned exclusively from partial normal training data $\mathcal{T}\subset\mathcal{N}$ without requiring explicit access to anomalous samples. This setting naturally motivates the model to assign a high likelihood to typical configurations drawn from $\mathcal{N}$, while deviations from this learned structure are identified as anomalies. This principle is realized by embedding the input features into a high-dimensional structured representation, enabling the systematic modeling of multivariate feature correlations under favorable scaling and optimization behavior. The schematic workflow of SMT-AD is shown in Fig.~\ref{fig: SMT_AD}.
\subsection{Preprocessing and feature embedding}
Before the training, the raw dataset is preprocessed to mitigate the influence of outliers and to ensure consistent feature scaling. Specifically, we apply a rank-based normalization independently to each feature. For a given feature $l$, the raw values are ordered, and each data point is mapped to a normalized value $\tilde{x}_{nl} = \mathsf{rank}_l(x_{nl})/N$ where $\mathsf{rank}_l$ denotes the rank of the raw data point $x_{nl}$ within feature $l$. This monotonic transformation suppresses the effect of extreme values and standardizes marginals to $\mathrm{Uniform}(0,1)$. For features that take discrete values, the normalization simplifies accordingly. If feature $l$ assumes $D_l$ distinct levels, the normalized representation can be written as $\tilde{x}_{nl} = \mathsf{rank}_l(x_{nl})/D_l$, which is consistent with the continuous rank normalization and preserves the ordering structure of the data.

As is well established in deep learning, introducing nonlinearity enhances a model's representational capacity and improves learning efficiency. Here, each normalized input vector $\vb*{\tilde{x}}_{n}$ is mapped to an input MPS, $\ket{\Psi_n}$, thereby enabling the model to capture nonlinear and multiscale correlations among features in a controlled manner. To further enrich the representation, we incorporate a frequency embedding, in which each input feature is mapped across multiple resolution scales with periodic structures. Accordingly, we define a feature map $\Psi: [0,1]^L \mapsto (\mathbb{R}^{2})^{\otimes PL}$, where the additional index $p=1,\ldots,P$ labels distinct frequency modes. In this work, we employ a Fourier-based embedding for each frequency mode $\omega_p := \pi/2^p$. For a fixed mode $p$, the corresponding input MPS is defined as
\begin{equation}
    \ket{\Psi_n^{(p)}} = \bigotimes_{l=1}^L \mqty(\cos\qty(\omega_p \tilde{x}_{nl} ) \\ \sin\qty(\omega_p \tilde{x}_{nl} )).
\end{equation}
By stacking multiple frequency modes, the full input representation $\ket{\Psi_n}$ encodes each feature across a hierarchy of frequencies, allowing the model to capture both coarse and fine-grained variations in the data.

\subsection{Matrix Product Operator}

After mapping each raw input $\vb*{x}_n$ to a nonlinear product-state feature MPS with Fourier embedding, we introduce a learnable but computationally light linear operator to increase expressivity without increasing the tensor-network bond dimension. Specifically, we utilize a constrained MPO built from sitewise $\mathsf{SO}(2)$ rotations and a superposition of $M$ mixture components across $P$ embedding resolutions. Here, $P$ indexes the resolution scale in the feature map, while $M$ controls the number of rank-1 MPO terms. Concretely, an $(m,p)$ component of the MPO at site $l$, defined as
\begin{equation}
    \mathsf{MPO}^{[l]}_{mp} = \mqty(\cos\theta^{mp}_l & -\sin\theta^{mp}_l \\ \sin\theta^{mp}_l & \cos\theta^{mp}_l),\qquad \theta^{mp}_l\in\mathbb{R}
\end{equation}
is applied to a $p$-element of the input MPS $\ket{\Psi_n^{(p)}}$ and we superpose all elements with coefficient $c_{mp}\in\mathbb{R}$, yielding an output MPS as
\begin{equation}
    \ket{\tilde{\Phi}_{n}} = \sum_{m=1}^M\sum_{p=1}^P c_{mp}\bigotimes_{l=1}^L \mqty(\cos\qty(\theta^{mp}_l + \frac{\pi}{2^p} \tilde{x}_{nl} ) \\ \sin\qty(\theta^{mp}_l + \frac{\pi}{2^p} \tilde{x}_{nl})),
\end{equation}
where $\Theta := \{c_{mp},\theta^{mp}_l\}$ are the MPO parameters. Note that this output MPS would be normalized by a normalization constant 
\begin{equation}
    \mathcal{Z}_n := \braket{\tilde{\Phi}_{n}} = \sum_{m,m' = 1}^M\sum_{p,p'=1}^P c_{mp}c_{m'p'} \prod_{l=1}^L \cos\qty(\theta^{mp}_l - \theta^{m'p'}_l + \qty(\frac{\pi}{2^p}-\frac{\pi}{2^{p'}}) \tilde{x}_{nl}),
\end{equation}
which depends on the data point.

There are multiple ways to turn a normalized output MPS $\ket{\Phi_n} = |\tilde\Phi_n\rangle/\sqrt{\mathcal{Z}_n}$ into a scalar prediction, for example by computing its overlap with a reference state or the expectation value of an observable. In this work, we use the squared overlap with a fixed reference state. Because our goal is anomaly detection, the reference state is chosen to represent ``normality'' and is set to the computational basis product state $\ket{0}^{\otimes L}$. We therefore define the \textit{normality score} 
\begin{equation}\label{eq: normality}
    a_\Theta(\vb*{x}_{n}) := \qty|\braket{0\cdots 0}{\Phi_n}|^2 = \frac{1}{\mathcal{Z}_n}\qty[\sum_{m=1}^M\sum_{p=1}^P c_{mp}\prod_{l=1}^L\cos\qty(\theta^{mp}_l+\frac{\pi}{2^p}\tilde{x}_{nl})]^2,
\end{equation}
which should be close to unity for normal data and significantly smaller for anomalous data.

\subsection{Training scheme}

Next, we train the model so that the embedded input MPS separates anomalous samples from normal ones. Concretely, for normal data, the output MPS should lie as close as possible to the reference state, which corresponds to maximizing the normality score. However, directly maximizing the normality score is numerically inconvenient because it is a product of $L$ cosine terms and therefore can typically become extremely small as $L$ grows. Thus, in the training, maximizing the normality score can alternatively be equivalent to minimizing the negative of a logarithm of the normality score (\textit{i.e.} the negative log-likelihood):
\begin{equation}
    \mathcal{L}_0(\Theta) = -\frac{1}{|\mathcal{T}|}\sum_{\vb*{x} \in \mathcal{T}} \log a_{\Theta}(\vb*{x}), 
\end{equation} 
where $\Theta = \Theta_c \cup \Theta_\theta$ and these sets are given by $\Theta_c = \{c_{mp}\}$ and $\Theta_\theta = \{\theta^{mp}_l\}$, and $|\mathcal{T}|$ is the size of the training data set. 
To stabilize the training and avoid parameter blow-up, we add regularization terms $\lambda_c$ and $\lambda_\theta$ that penalize large coefficients in $\Theta$ with Tikhonov regularization    
\begin{equation} 
\mathcal{R}(\Theta) = \lambda_c\norm{\Theta_c}^2_F+ \lambda_\theta\norm{\Theta_\theta}^2_F. 
\end{equation}
The $\lambda_c$ and $\lambda_\theta$ are thus regularization hyperparameters for MPO's parameter sets $\Theta_c$ and $\Theta_\theta$, respectively, and we optimize their value to obtain better results. Therefore, the final optimization loss is $\mathcal{L} = \mathcal{L}_0(\Theta) + \mathcal{R}(\Theta)$. After training, we denote the score produced by the optimal parameters $\Theta^*$ as $a(\vb*{x}):=a_{\Theta^*}(\vb*{x})$.

\section{Numerical Experiments}\label{Section: Experiments}
In our numerical experiment, we use the \textit{Wine}, \textit{Lymphography}, \textit{Thyroid}, and \textit{Satellite} datasets from the UCI repository \cite{Dua:2019}, together with the \textit{Credit Card} dataset from Kaggle \cite{creditcardfraud_kaggle_2013}. Among these, only the \textit{Credit Card} data are preprocessed with the principal component analysis (PCA) prior the anomaly detection; the remaining datasets are used in their original (raw) form. The number of data points in each dataset are shown in Table~\ref{table: dataset}.

Because several of these datasets are multiclass, we follow Ref.~\cite{wang2020anomaly} and designate a subset of classes as normal data $\mathcal{N}$ and treat the remaining classes as anomalies $\mathcal{A}=\mathcal{D}\setminus\mathcal{N}$. After preprocessing, we randomly split half of the normal dataset as a training dataset $\mathcal{T}$ and use the remaining normal samples $\mathcal{N}\setminus\mathcal{T}$ and all anomalous samples $\mathcal{A}$ for testing. Model parameters $\Theta$ are learned using mini-batch optimization, updating sequentially over batches. We evaluate anomaly-detection performance with threshold-independent metrics: the area under the receiver operating characteristic curve (AUROC) and the area under the precision-recall curve (AUPRC). The AUROC quantifies the probability that an anomalous sample receives a lower normality score than a normal one (with 0.5 corresponding to random guessing and 1 to perfect ranking), while the AUPRC summarizes the trade-off between precision and recall across all decision boundaries and is better suited to the high-imbalance anomaly detection task.

\begin{table}[]
\caption{Information of datasets, sorted by size.}
\label{table: dataset}
\begin{tabular}{|c||c|cc|c|}
\hline
\multirow{2}{*}{Dataset} & \multirow{2}{*}{\#Training $|\mathcal{T}|$} & \multicolumn{2}{c|}{\#Test} & \multirow{2}{*}{\#Feature $L$}                                                                                                \\ \cline{3-4} 
                         &    & \multicolumn{1}{c|}{Normal} & Anomalous & \\ \hline
\textit{Wine}                    &  59          & \multicolumn{1}{c|}{60 (85.7\%)} &   \multicolumn{1}{c|}{10 (14.3\%)} &  13          \\ 
\textit{Lympho}                   &  71          & \multicolumn{1}{c|}{71 (92.2\%)} &   \multicolumn{1}{c|}{6 (7.8 \%)} & 18           \\ 
\textit{Thyroid}                  &  1839        & \multicolumn{1}{c|}{1840 (95.2\%)} & \multicolumn{1}{c|}{93 (4.8\%)}  & 6             \\ 
\textit{Satellite}                &  2199        & \multicolumn{1}{c|}{2200 (51.9\%)} & \multicolumn{1}{c|}{2036 (48.1\%)} & 36              \\ 
\textit{Credit Card}              &  142403          & \multicolumn{1}{c|}{142404 (99.83\%)} &  \multicolumn{1}{c|}{492 (0.17\%)} & 30             \\ \hline
\end{tabular}
\end{table}

\subsection{Implementation}\label{section: implement}
In the numerical experiments, the baseline models, \textit{i.e.}, one-class support vector machine (OC-SVM) and isolation forest (IF) were implemented by Scikit-learn library, and TNAD \cite{wang2020anomaly} and SMT-AD  were implemented by PyTorch library to leverage GPU acceleration, with AdamW used as the optimizer. The AUROC and AUPRC were calculated using the Scikit-learn library. The AUROC and AUPRC are reported as the best mean $\pm$ standard deviation across internal parameters and hyperparameters grids over 20 realizations of initial parameters and randomly selected training data $\mathcal{T}$.

For the baseline models, we follow Ref.~\cite{wang2020anomaly} to utilize the hyperparameters search. For all OC-SVM numerical experiments, the radial basis function kernel was used, and a grid sweep was conducted for the kernel coefficient $\gamma \in \{2^{-10},\ldots, 2^{-1}\}$ and the margin parameter $\nu=\{0.01,0.1\}$. For all IF numerical experiments, the number of trees and the sub-sampling size $|\mathcal{B}|$ were set to 100 and 256, respectively, as recommended by the original paper \cite{liu2008isolation}.

For TNAD, we also follow Ref.~\cite{wang2020anomaly} by setting the bond dimension of the MPO $\chi=5$ for all numerical experiments, batch size $|\mathcal{B}|$ is 64 for small datasets, \textit{i.e.}, \textit{Wine}, \textit{Lymphography}, and \textit{Thyroid}, and 512 for large datasets, \textit{i.e.}, \textit{Satellite} and \textit{Credit Card}, moreover, training epoch $T_\text{epoch}$ is 40 for all experiments. For simplicity, the spacing of the MPO's output legs $S$ is equal to 1. We perform the grid sweep for the number of Fourier terms $P\in\{2,4,6,8\}$, and the regularization parameter $\alpha$ and learning rate $\eta$ are $(\alpha,\eta) = (0.1,1.0\times10^{-3})$ for \textit{Wine}, \textit{Lymphography}, \textit{Thyroid} and $(\alpha,\eta) = (0.3,5.0\times10^{-4})$ for \textit{Satellite} and \textit{Credit Card} datasets.

The details of SMT-AD's internal parameters are as follows: constant learning rate $\eta = 0.01$ during the training, batch size $|\mathcal{B}|$ is 64 for small dataset, \textit{i.e.}, \textit{Wine}, \textit{Lymphography}, and \textit{Thyroid}, and 512 for large dataset, \textit{i.e.}, \textit{Satellite} and \textit{Credit Card}, training epoch $T_\text{epoch}$ is determined based on the number of training data as $T_\text{epoch} = \lfloor 15000 |\mathcal{B}|/|\mathcal{T}|\rfloor$, which is determined heuristically based on convergence behavior, and we fix regularization parameters $\lambda_c = 0.01$ and $\lambda_\theta = 0.001$. In the best performance search, we perform a grid search in $M\in \{2,4,6,\ldots,40\}$ and $P\in\{1,2,3,4\}$.

In our experiments, SMT-AD and TNAD are constructed and optimized within the PyTorch framework with GPU acceleration, while OC-SVM and IF are performed by Scikit-Learn and NumPy frameworks. All SMT-AD and TNAD are executed on nodes equipped with NVIDIA A100 Tensor Core GPUs, and CPUs AMD EPYC$^\text{TM}$ 7713 processors are used to process all OC-SVM and IF \cite{NSCC}.

\subsection{Results}
Table \ref{table: AUROC} summarizes anomaly-detection performance across datasets, reported as the mean AUROC and AUPRC ($\pm$ standard deviation) over 20 realizations. Overall, SMT-AD achieves consistently strong AUROC, matching or exceeding OC-SVM, IF, or even TNAD on all five benchmarks. In particular, SMT-AD attains near-ceiling AUROC on \textit{Wine}, \textit{Lymphography}, and \textit{Thyroid}, and remains competitive on the more challenging \textit{Satellite} dataset. Note that the standard deviation is less than 0.05\%, so we then report only 0.1\% in the table. The AUPRC results largely follow the same trend---SMT-AD is comparable to the strongest baselines on most datasets, indicating good precision-recall behavior under imbalance. The main exception is the \textit{Credit Card} dataset, where SMT-AD retains the highest AUROC but exhibits a markedly lower AUPRC than OC-SVM and TNAD, suggesting that while anomalies are ranked higher on average (signalled by large AUROC), the detection threshold suffers from increased false positive (signalled by a lower AUPRC). 
This said, since the \textit{Credit Card} dataset is highly imbalanced, whereby anomalous data only correspond to 0.17\% of the total, even a AUPRC of about 38\% corresponds as we get from SMT-AD, corresponds to a 200-fold improvement on detection from an untrained scenario.  

\begin{figure}
    \centering
    \includegraphics[width=0.8\linewidth]{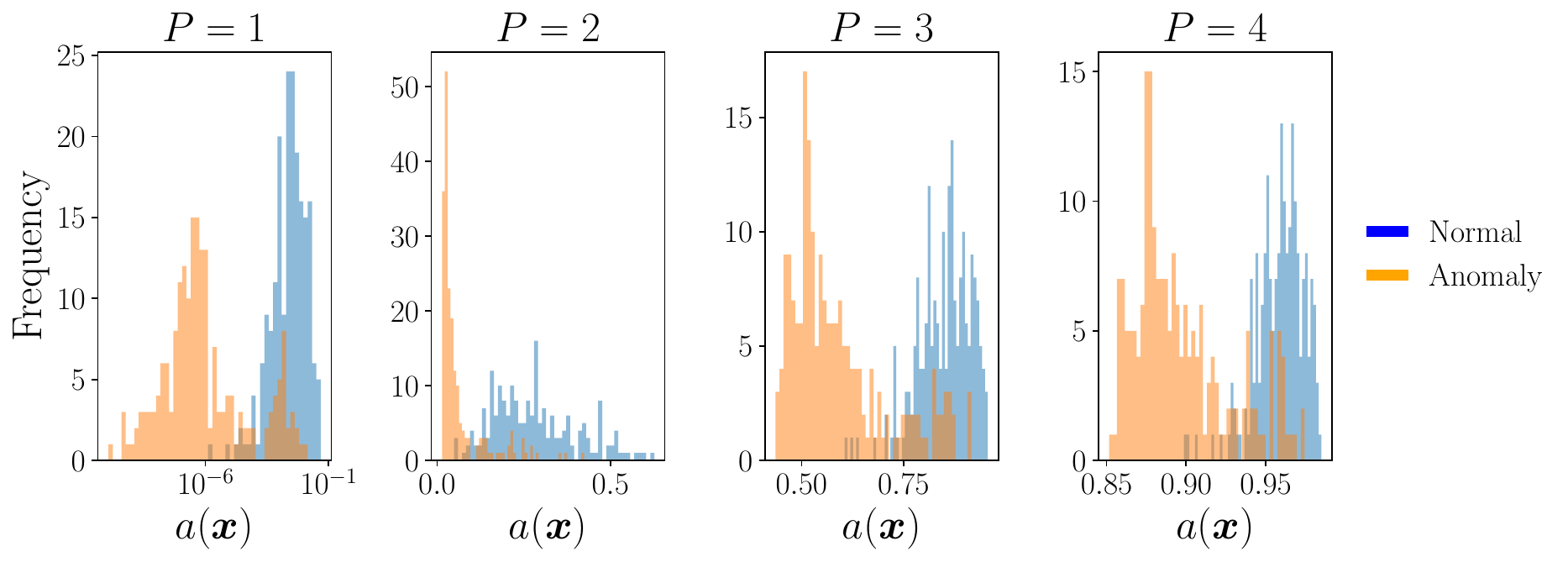}
    \caption{Distributions of normality score for 200 normal and 200 anomalous \textit{Credit Card} samples. Results are shown across varying embedding resolutions $P\in\{1,2,3,4\}$ for a trained model with $M=30$.}
    \label{fig: distribution}
\end{figure}

Focusing on \textit{Credit Card} dataset, Fig.~\ref{fig: distribution} plots histograms of the normality score $a(\vb*x)$ for 200 normal and 200 anomalous samples under different embedding resolutions $P$ with $M=30$. For $P=1$, scores concentrate at extremely small values (near $10^{-6}$), whereas for $P=4$ they collapse toward values close to one. These two extremes indicated under- and over-confident mappings, respectively, both of which reduce effective score contrast. Intermediate resolutions $P=2$ and $P=3$ yield better-calibrated distributions---the scores spread over a wider dynamic range, and the separation between normal and anomalous histograms becomes more apparent, especially for $P=2$. The left panel of Fig.~\ref{fig: creditcard_roc_prc} confirms this finding that $P=2$ has the best AUROC and AUPRC at $M=30$. Additionally, we find that AUROC and AUPRC increase when $M$ increases and saturate at a certain value of $M$ (in the plot, AUPRC is saturated at $M\sim 16$) for $P>1$ and continue to increase for $P=1$. Therefore, for large enough $M$, $P$ acts as a calibration parameter such that overly small or large $P$ compresses the score distribution and harms discrimination, while intermediate values of $P$ yield a better-separated normality score for anomaly detection.

\begin{table}[]
\caption{Average ($\pm$ standard deviation) AUROC and AUPRC in anomaly detection task from several baseline models. These results are averaged over 20 realizations.}
\label{table: AUROC}
\begin{tabular}{|c||cccc||cccc|}
\hline
\multirow{2}{*}{Dataset}           & \multicolumn{4}{c||}{AUROC}                                                                                                        & \multicolumn{4}{c|}{AUPRC}                                                                                                        \\ \cline{2-9}
    & \multicolumn{1}{c|}{OC-SVM}         & \multicolumn{1}{c|}{IF}             & \multicolumn{1}{c|}{TNAD}            & SMT-AD         & \multicolumn{1}{c|}{OC-SVM}         & \multicolumn{1}{c|}{IF}             & \multicolumn{1}{c|}{TNAD}            & SMT-AD         \\ \hline
\textit{Wine}      & \multicolumn{1}{c|}{$98.1 \pm 1.1$} & \multicolumn{1}{c|}{$99.0 \pm 0.6$} & \multicolumn{1}{c|}{$97.6 \pm 1.0$} & $98.4 \pm 0.1$ & \multicolumn{1}{c|}{$97.3 \pm 1.8$} & \multicolumn{1}{c|}{$98.3 \pm 1.4$} & \multicolumn{1}{c|}{$95.9 \pm 1.9$} & $97.6 \pm 0.1$          \\ 
\textit{Lympho}     & \multicolumn{1}{c|}{$99.9 \pm 0.1$} & \multicolumn{1}{c|}{$97.9 \pm 1.6$} & \multicolumn{1}{c|}{$99.3 \pm 0.8$}  & $99.8 \pm 0.1$ & \multicolumn{1}{c|}{$99.2 \pm 1.6$} & \multicolumn{1}{c|}{$85.5 \pm 8.4$} & \multicolumn{1}{c|}{$93.8 \pm 6.5$} & $98.4\pm0.1$               \\ 
\textit{Thyroid}    & \multicolumn{1}{c|}{$97.0 \pm 0.5$} & \multicolumn{1}{c|}{$96.9 \pm 1.0$} & \multicolumn{1}{c|}{$98.5 \pm 0.3$}                &  $99.1 \pm 0.1$              & \multicolumn{1}{c|}{$57.3 \pm 5.0$} & \multicolumn{1}{c|}{$60.3 \pm 9.1$} & \multicolumn{1}{c|}{$61.5 \pm 9.6$} & $69.3 \pm 0.6$       \\ 
\textit{Satellite}  & \multicolumn{1}{c|}{$68.1 \pm 0.3$} & \multicolumn{1}{c|}{$78.0 \pm 1.2$} & \multicolumn{1}{c|}{$79.8\pm1.3$}                & $75.9\pm0.1$               & \multicolumn{1}{c|}{$78.7 \pm 0.2$} & \multicolumn{1}{c|}{$83.2 \pm 0.7$} & \multicolumn{1}{c|}{$84.7\pm0.9$}                &  $81.7\pm0.1$              \\ 
\textit{Credit Card} & \multicolumn{1}{c|}{$93.9 \pm 0.2$}               & \multicolumn{1}{c|}{$94.3 \pm 0.3$}               & \multicolumn{1}{c|}{$92.0 \pm 0.4$}  & $94.8 \pm 0.1$ & \multicolumn{1}{c|}{$64.0 \pm 2.2$}               & \multicolumn{1}{c|}{$29.1 \pm 5.7$}               & \multicolumn{1}{c|}{$72.7 \pm 1.7$}  & $36.9 \pm 0.1$ \\ \hline
\end{tabular}
\end{table}

\section{Analysis}\label{section: analysis}

The results show that SMT-AD can achieve strong anomaly detection performance with high computational efficiency. In this section, we analyze \textit{how} the method works by examining how its two key hyperparameters $(P,M)$ control the expressivity of the model and, consequently, the separability between normal and anomalous samples. Increasing $P$ enriches the local nonlinear embedding at each site, while increasing $M$ enlarges the space of superposed MPO terms available during training. Moreover, we focus our interpretability analysis on $P$ showing that increasing $P$ changes the structure captured by the model via the feature importance analysis and the feature-feature correlation analysis.

\subsection{Feature importance via entanglement entropy}\label{subsec: FI}

In many real-world datasets, anomalies are characterized not only by unusual feature magnitudes but also by changes in cross-feature-dependency structure or correlations. Since our model is a quantum-inspired model, dependencies are naturally reflected by entanglement. If a feature (site) is weakly coupled to the rest of all features (chain), its one-site-reduced state remains nearly pure, and the corresponding entropy is small; conversely, a large single-site entropy indicates that information at that site is distributed nonlocally through correlations with other features.

\begin{figure}
    \centering
    \includegraphics[width=1.0\linewidth]{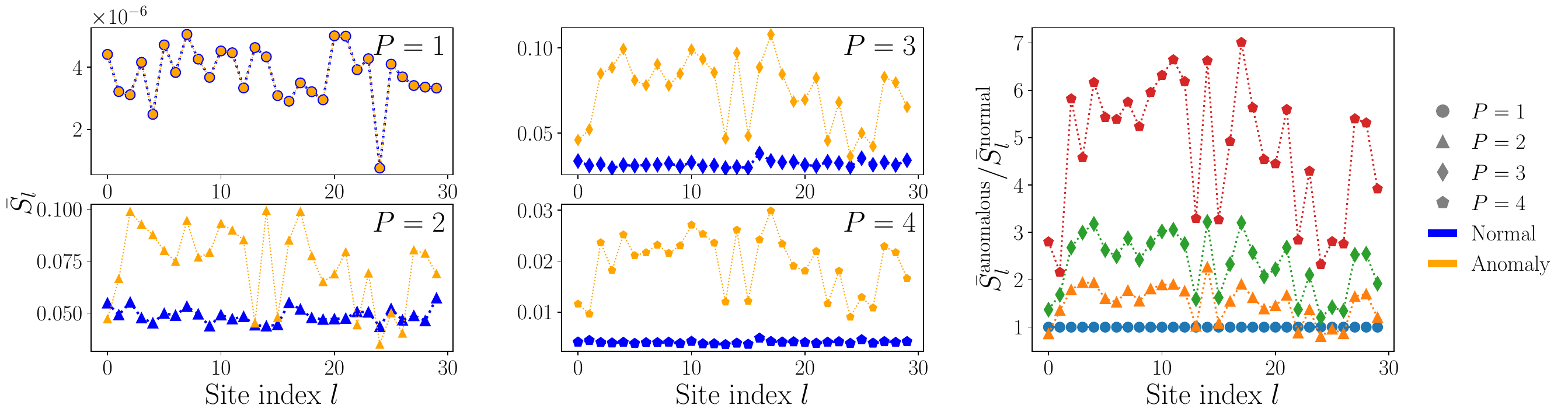}
    \caption{Feature importance analysis via single-site entanglement entropy for the \textit{Credit Card} dataset from the trained model with $M=30$. (Left and middle) The averaged single-site entanglement entropy from $P=1$ to $P=4$ and (right) the amplification ratio of anomalous to normal entanglement entropy $\bar{S}_l^{\text{anomalous}} / \bar{S}_l^{\text{normal}}$ across all features for 200 normal and 200 anomalous samples.}
    \label{fig: feature_importance}
\end{figure}

\begin{figure}
    \centering
    \includegraphics[width=0.7\linewidth]{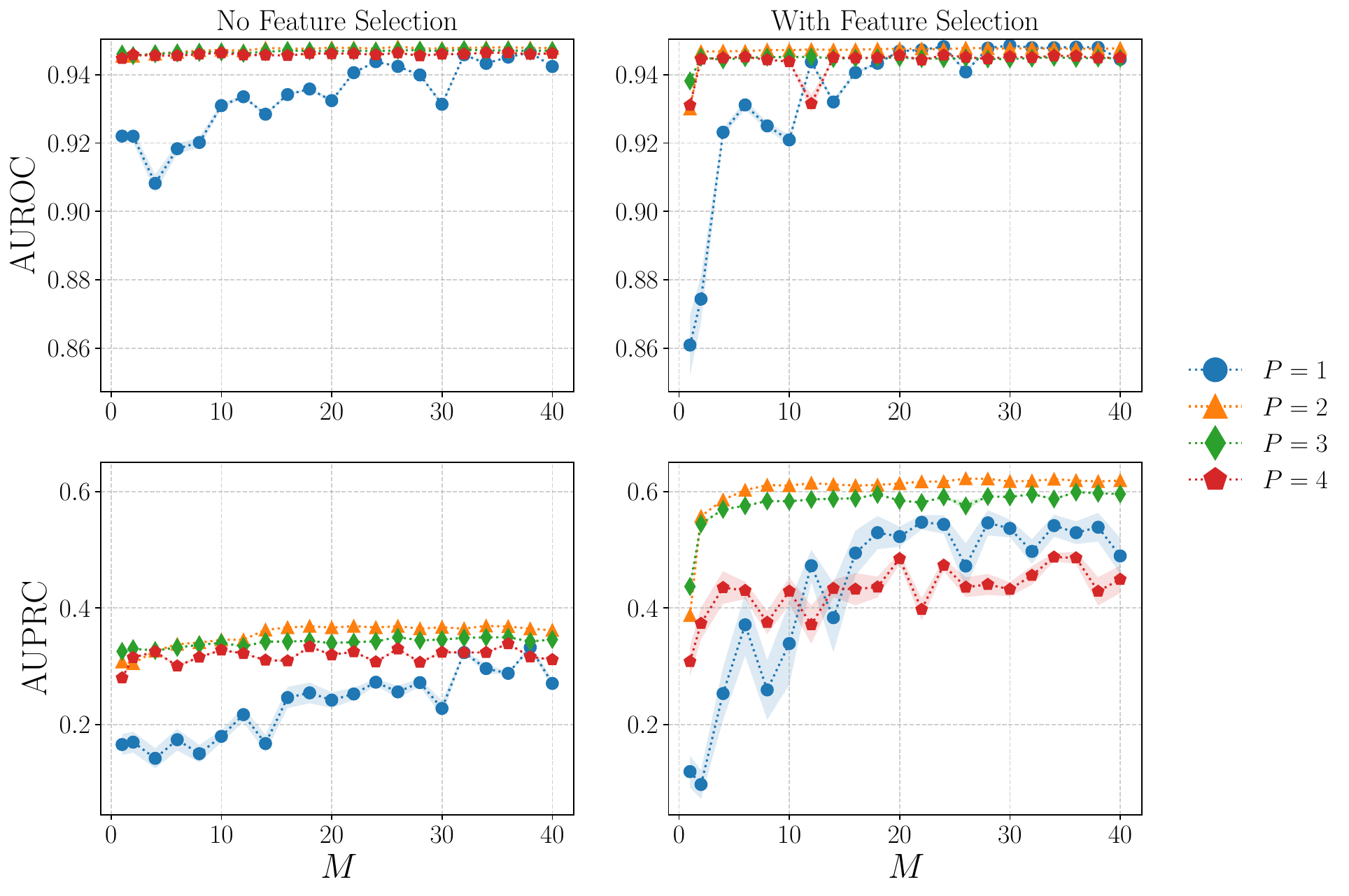}
    \caption{Comparative performance analysis of AUROC and AUPRC across varying $M$ from 1 to 40 for the \textit{Credit Card} dataset, and embedding resolution $P$. The left column shows results without feature selection, while the right column displays results with feature selection applied. Performance is measured using (top row) AUROC and (bottom row) AUPRC for values of $P$ ranging from 1 to 4. Shaded areas represent the standard deviation intervals across 20 numerical experimental trials.}
    \label{fig: creditcard_roc_prc}
\end{figure}

To quantify how strongly the trained model couples information across the embedded feature chains, we consider the dataset-averaged single-site entanglement entropy from the trained output MPS at each feature $l$, $\bar{S}_l = \mathbb{E}_n\qty[S_l\qty(\ket{\Phi_n})]$, for varying model parameter $P$. The single-site entanglement entropy at site $l$ can be computed from $S_l(\ket{\Phi}) = -\Tr{\rho_{l}\ln\rho_{l}}$ where $\rho_l = \Tr_{\setminus \{l\}} \ketbra{\Phi}$ is the site-$l$ reduced density matrix. The left and central panels in Fig.~\ref{fig: feature_importance} compare the averaged entropy profiles of 200 normal (blue) against 200 anomalous (orange) \textit{Credit Card} samples for $P=1$ to $P=4$. For $P=1$, the entropy contributions are negligible and indistinguishable between normal and anomalous samples, whereas for $P>1$ the profiles become clearly separable, indicating that the richer Fourier embedding and the superposed MPO activate class-dependent nonlocal structure in the output MPS. Interestingly, the anomaly detection performs well even with $P=1$, as shown in Fig.~\ref{fig: creditcard_roc_prc} (left panels). However, the emergence of significant entanglement entropy for $P>1$ reveals that the model begins to capture the subtle non-linear dependencies that are not grasped for $P=1$. 
To quantify this structural deviation, we analyze the entanglement entropy amplification ratio $\bar{S}_l^{\text{anomalous}} / \bar{S}_l^{\text{normal}}$ (right panel in Fig.~\ref{fig: feature_importance}). While this ratio remains near unity at $P=1$, it rises sharply to range between $2.5$ and $6.0$ at $P=4$. This amplification acts as a local sensitivity metric---the peaks highlight the latent dimensions where the anomaly most strictly deviates from the learned nonlinear manifold. 

Finally, we leverage these high-entropy signatures for feature selection to validate their importance. By selecting only the features that exhibit high entanglement entropy in the anomalous samples, we retrain the model by including only those features and re-evaluate detection performance. Here, we select features at site indexes $2-12,14,16-18,21,27,\text{and } 28$.

As illustrated in the right panel of Fig.~\ref{fig: creditcard_roc_prc} compared with the left panel, while the AUROC remains fairly constant, the AUPRC increases significantly. Additionally, for  $P=2$ and $P=3$, the AUROC/AUPRC saturate with lower number of $M$ compared with no selection case (AUROC/AUPRC saturates at $M\sim10$). This trade-off indicates that the high-entropy features encapsulate the most critical information, thereby improving the precision of the detection and the overall training efficiency in the imbalanced regime. Moreover, we can see the stability in the performance for $P=2$ and $P=3$, while there is no improvement in the performance for $P=4$. This indicates that the model with $P=4$ has already learned high-entropy features during training, even when we train the model without the feature selection. Table \ref{table: feature_selection} shows the best performances of SMT-AD considering feature selection, with significant improvement compared with SMT-AD with full feature training and comparable with TNAD's performances. 

At this point, it is important to try to better understand the performance of SMT-AD with respect to TNAD. In a recent work, \cite{SornsaengPoletti2026}, we have studied the performance of an ansatz analogous to SMT-AD, called the superposition of product states (SPS), to represent the ground state of the quantum Ising model, and compared it to the use of matrix product states (MPS) which, instead, is analogous to TNAD. In \cite{SornsaengPoletti2026}, we found that SPS could describe very accurately ground states in the ferromagnetic phase, but it was not able to describe accurately ground states in the paramagnetic phase, which are low-correlated ground states in a disordered background, showing some limitations in the expressive power. The scenario here is analogous---while SMT-AD is well described on correlated datasets, it may struggle when the dataset has low correlations. To show this more clearly, we have conducted a numerical experiment using, as a data set, spin configurations drawn from ground states of the transverse-field Ising model, which corresponds to a vector of 0s or 1s. We considered two types of ground state from which to sample configurations, near-critical (highly correlated) and paramagnetic (uncorrelated) ground states (detailed in App. \ref{app: phase}). 
We observe, as expected, that while SMT-AD excels at mapping highly correlated spin configurations drawn from the near-critical ground state, its performance degrades when confronted with uncorrelated noise characteristic of a paramagnetic regime. This behavior mirrors the difficulty of SMT-AD on the PCA-orthogonalized \textit{Credit Card} dataset, where SMT-AD underperforms relative to an unconstrained baseline like TNAD.  

\subsection{Feature-Feature correlation}

We now examine the feature-feature correlation of the model to understand how the model differently correlates the features for normal and anomalous data. To quantify this, we utilize the pairwise mutual information (MI) measure, computed from the entanglement entropy of the trained output MPS. Although the input features are linearly decorrelated, MI exposes the non-separable interactions that the trained model induces in its latent representation. Concretely, the feature-feature correlation between features $k$ and $l$ from any data point $\vb*{x}\in \mathcal{D}$ (encoded as $\ket{\Phi}$) is utilized by
\begin{equation}
    I_{k,l}(\ket{\Phi}) = S_k(\ket{\Phi}) + S_l(\ket{\Phi}) - S_{k,l}(\ket{\Phi})
\end{equation}
where $S_k$ is the entanglement entropy at site $k$, and $S_{k,l}$ is the two-site entanglement entropy at sites $k$ and $l$, computed from $S_{k,l} = -\Tr{\rho_{k,l}\ln\rho_{k,l}}$ where $\rho_{k,l} = \Tr_{\setminus \{k,l\}} \ketbra{\Phi}$. We report the dataset-averaged MI matrices $\bar{I}_{k,l} = \mathbb{E}_n[I_{k,l}(\ket{\Phi_n})]$ for 200 normal and 200 anomalous subsets.

\begin{figure}
    \centering
    \includegraphics[width=1.0\linewidth]{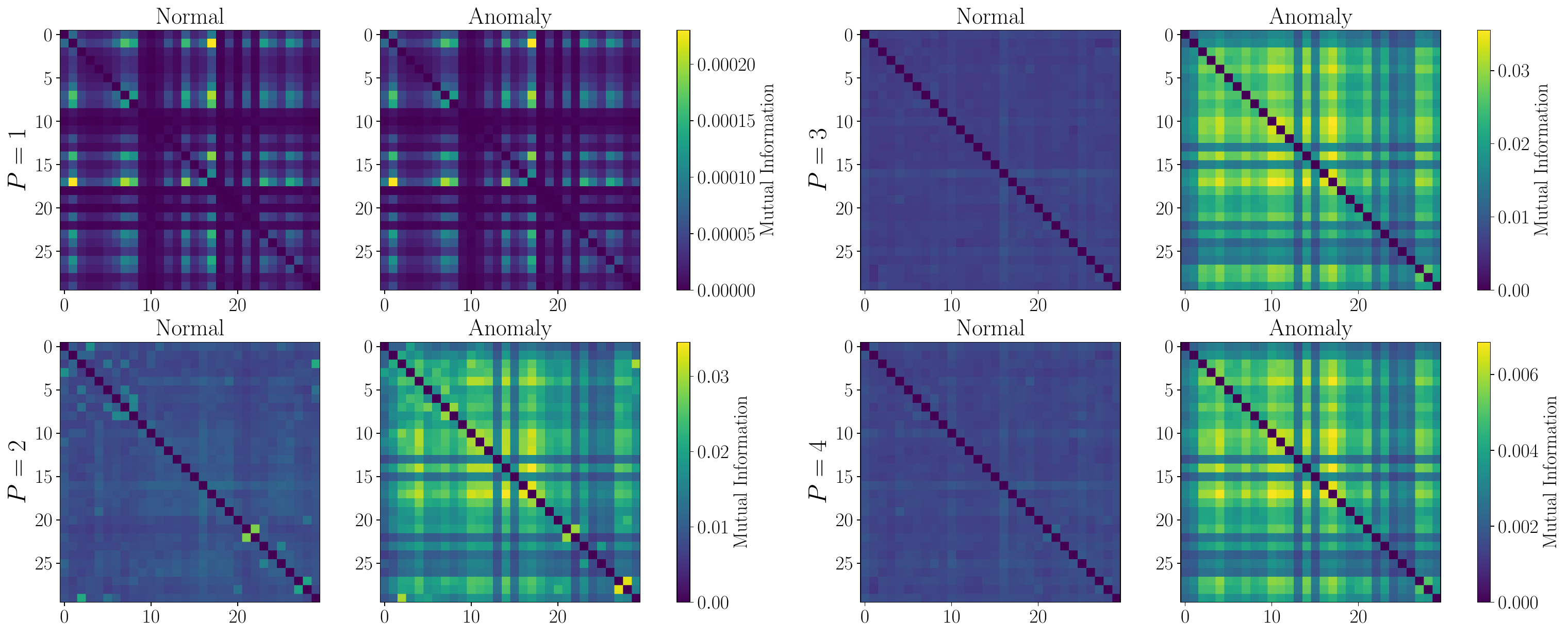}
    \caption{Average mutual information matrix of the trained model over 200 normal and 200 anomalous samples for varying $P$ from the trained model with $M=30$ for the \textit{Credit Card} dataset.}
    \label{fig: MI}
\end{figure}

Figure \ref{fig: MI} shows average MI matrices for dataset $\mathcal{N}$ and $\mathcal{A}$ across embedding resolution $P$. A clear
transition occurs between $P=1$ and $P>1$. For $P=1$, as seen in the same situation in Sec.~\ref {subsec: FI}, the average MI matrices for both normal and anomalous data are exactly the same, and the magnitudes are nearly identical and remain close to zero, indicating that the learned representation is largely factorized and weakly dependent on the data distribution. In contrast, for $P>1$, normal samples maintain weak and diffuse MI, consistent with normal data lying on a low-entanglement manifold, \textit{i.e.} the target state is a product state. Meanwhile, the anomalous set exhibits substantially larger MI with distinct structured patterns, where certain features behave as interaction hubs that correlate. This shows a clear separation between anomalous and normal data, whereby anomalies are characterized by a collective reorganization of feature-feature correlations, rather than by localized deviations of single features.

\begin{table}[]
    \caption{Performance comparison of SMT-AD using full features versus entanglement-assisted feature selection.}
    \label{table: feature_selection}
    \centering
    \begin{tabular}{|c||c|c|}
    \hline
         & AUROC & AUPRC \\
    \hline
       Full features  & $94.8 \pm 0.1$ & $36.9 \pm 0.1$     \\

       Features selection & $94.8 \pm 0.1$  & $62.2 \pm 0.1$     \\
    \hline
    \end{tabular}
    
\end{table}

\begin{table}[]
\caption{Number of parameters, time complexities, optimal hyperparameters with their corresponding number of internal parameters for achieving an AUPRC (comparable with the mean value shown in Table \ref{table: AUROC}), and processing times (training and inference per sample) across benchmarks in the anomaly detection for the \textit{Credit Card} dataset. Time complexity with an asterisk ($^*$) is for one training epoch.}
\label{table: complexity}
\begin{tabular}{|c||c|c||c|c|c|c|}
\hline
Model  & \#Parameter                  & Time ($^*$per epoch)   & Hyperparameters   & \#Parameter & Training time (s) & Infer time ($\mu$s)     \\ \hline
OC-SVM & $N_\text{sv}L+N_\text{sv}+1$ & $O(|\mathcal{T}|^2L + |\mathcal{T}|^3)$ & $N_\text{sv}=1454$ & 45075 & 35.83 $\pm$ 53.86 & 115.64 $\pm$ 155.21                                \\ 
IF     & $N_\text{tree}|\mathcal{B}|$ & (expected) $O(N_\text{tree}|\mathcal{B}|\log |\mathcal{B}|)$ & $N_\text{tree} = 100$ & 20348 & 0.21 $\pm$ 0.01 & 4.29 $\pm$ 0.06 \\ 
TNAD   & $L\chi^2P^2$                 & $O(L\chi^2(\chi+P)(P+1)|\mathcal{B}|)^*$ & $\chi = 4, P = 6$ & 17280 & 1415.28 $\pm$ 142.51 & 319.71 $\pm$ 35.00         \\ 
SMT-AD & $MP(L+1)$                    & $O(LMP(MP+1)|\mathcal{B}|)^*$     & $M = 10, P = 2$ & 620 & 864.48 $\pm$ 109.29 & 45.41 $\pm$ 7.25       \\ \hline
\end{tabular}
\end{table}
\subsection{Computational complexities}
Table~\ref{table: complexity} summarizes the number of learnable parameters and the time complexity of each baseline, using the additional notation: $N_\text{sv}$ is the number of support vectors for OC-SVM, $N_\text{tree}$ is the number of trees for IF, $|\mathcal{B}|$ is the batch size (for TNAD and SMT-AD) or the sub-sampling size (for IF), and $\chi$ is the learnable MPO bond dimension for TNAD. 

For OC-SVM, the model stores $N_\text{sv}$ support vectors in $\mathbb{R}^L$ together with coefficients and a bias parameter, giving a parameter count of $N_\text{sv}L+N_\text{sv}+1$. The training cost is dominated by kernel matrix calculation and quadratic-program optimization, scaling as $O(|\mathcal{T}|^2L + |\mathcal{T}|^3)$, which can become prohibitive for large training data set size $|\mathcal{T}|$. 

For IF, the effective model size scales with the random isolation trees $N_\text{tree}$ and the sub-sampling size $|\mathcal{B}|$. Since each tree is grown by recursive random partitioning with expected depth $O(\log|\mathcal{B}|)$, the expected total training cost scales as $O(N_\text{tree}|\mathcal{B}|\log |\mathcal{B}|)$.

While TNAD's parameters count scales as $L\chi^2P^2$, Ref.~\cite{wang2020anomaly} reports that contracting an input MPS with a learnable MPO during training requires $O(L\chi^2(\chi+P)(P+1)|\mathcal{B}|)$ operations per training epoch (marked by $^*$).

Finally, SMT-AD has a compact parameterization $MP(L+1)$, which grows linearly with the number of features and MPO hyperparameter $(M,P)$. In the loss function evaluation, the numerator of the normality score \eqref{eq: normality} can be computed with $O(LMP)$ operations, whereas computing the normalization constant $\mathcal{Z}_n$ takes $O(LM^2P^2)$ operations. Consequently, the overall loss function computation scales as $O(LMP(MP+1)|\mathcal{B}|)$ per epoch per batch. Although the time complexity is comparable to TNAD, SMT-AD is considerably more parallelization-friendly in practice. The per-site contractions can be broadcast over both the batch and the $(M,P)$ channels, whereas TNAD typically requires sweep-wise left/right environment propagation and local tensor updates that proceed sequentially along the MPS chain, limiting effective parallelism. To illustrate this efficiency in practice, Table \ref{table: complexity} also shows that SMT-AD achieves optimal performance on the \textit{Credit Card} dataset with merely 620 parameters---orders of magnitude fewer than other baselines. The number of parameters is reduced even further to 380 after feature selection, while improving the performance. As for the computational time, we note that IF performs the fastest on both training and inference (for the number of parameters corresponding to its optimal performance), while SMT-AD is the second fastest among the models considered.

\section{Conclusion}\label{section: conclusion}

SMT-AD presents a highly scalable, tensor-network-inspired framework for anomaly detection. By mapping rank-normalized input data into a product-state MPS via Fourier-assisted multiresolution embedding, the model processes data through a superposition of bond-dimension-one learnable MPOs and exclusively learns a reference manifold from purely normal training data. Notably, its parameter count scales linearly with the feature size, the number of Fourier embedding resolution $P$, and the number of MPO components $M$. Furthermore, computing the normality score and loss function scales quadratically with $M$ and $P$, effectively bypassing the prohibitive cubic complexity associated with dataset size seen in OC-SVM. Across tabular benchmarks, SMT-AD consistently achieves AUROC scores that match or surpass the established baselines such as OC-SVM, IF, and TNAD, and similarly for AUPRC on almost all data sets. In particular, an intermediate embedding resolution (such as $P=2$ or $P=3$) with a small $M$, SMT-AD achieves anomaly detection performance on par with existing anomaly detection baseline methods.

Fundamentally, SMT-AD captures feature importance and feature-feature correlations through its embedding resolution and superposition, as demonstrated by single-site entanglement entropy and mutual information matrices. These entropy signatures identify important features and visualize the complex feature-feature correlations that separate anomalies from normal data, directly contributing to enhanced detection precision. When allowing the use of information from anomalous data, the integration of entanglement-guided feature selection directly into the SMT-AD protocol during training will automatically identify and remove low-entropy, uninformative features during the training, achieving end-to-end robustness against orthogonalized noise and improving the detection performance coherently. 
Additionally, while it is not as expressive as TNAD, its highly parallelizable, vectorizable, and scalable computational structure allows the algorithm to run efficiently even on low-end computing systems. This low-resource footprint makes SMT-AD a promising candidate for deployment in edge computing and internet-of-things environments.

\section*{Acknowledgment} 
D.P. and A.S. acknowledge the support of the Ministry of Education, Singapore, under the grant T2EP50123-0017, and from HTX under project HTX000ECI24000267. D.P. and A.S. acknowledge fruitful discussions with De Wen Soh. The authors also acknowledge fruitful discussions with Martin Trappe. The computational work was performed at the National Supercomputing Centre, Singapore \cite{NSCC}.

\section*{Data availability}

The raw data required to reproduce the above findings are available to download from the UCI repository for \textit{Wine}, \textit{Lymphography}, \textit{Thyroid}, and \textit{Satellite} datasets \cite{Dua:2019} and from Kaggle for \textit{Credit Card} dataset \cite{creditcardfraud_kaggle_2013}. The source code supporting these findings is publicly available \cite{GITHUB}.

\bibliography{ref}

@misc{NSCC,
  title = {http://nscc.sg},
  url = {http://nscc.sg}
}

@misc{GITHUB,
  title = {https://github.com/SUTD-MDQS/SMT-AD},
  url = {https://github.com/SUTD-MDQS/SMT-AD}
}

@article{pang2021deep,
  title={Deep learning for anomaly detection: A review},
  author={Pang, Guansong and Shen, Chunhua and Cao, Longbing and Hengel, Anton Van Den},
  journal={ACM computing surveys (CSUR)},
  volume={54},
  number={2},
  pages={1--38},
  year={2021},
  publisher={ACM New York, NY, USA},
  url={https://dl.acm.org/doi/abs/10.1145/3439950}
}

@article{aizpurua2025tensor,
  title={Tensor networks for explainable machine learning in cybersecurity},
  author={Aizpurua, Borja and Palmer, Samuel and Orus, Roman},
  journal={Neurocomputing},
  pages={130211},
  year={2025},
  publisher={Elsevier},
  url={https://www.sciencedirect.com/science/article/pii/S0925231225008835}
}

@article{efthymiou2019tensornetwork,
  title={TensorNetwork for Machine Learning},
  author={Efthymiou, Stavros and Hidary, Jack and Leichenauer, Stefan},
  journal={arXiv preprint arXiv:1906.06329},
  year={2019},
  url={https://arxiv.org/abs/1906.06329}
}

@article{wang2020anomaly,
  title={Anomaly detection with tensor networks},
  author={Wang, Jinhui and Roberts, Chase and Vidal, Guifre and Leichenauer, Stefan},
  journal={arXiv preprint arXiv:2006.02516},
  year={2020},
  url={https://arxiv.org/abs/2006.02516}
}

@article{stoudenmire2016supervised,
  title={Supervised learning with tensor networks},
  author={Stoudenmire, Edwin and Schwab, David J},
  journal={Advances in neural information processing systems},
  volume={29},
  year={2016},
  url={https://proceedings.neurips.cc/paper/2016/hash/5314b9674c86e3f9d1ba25ef9bb32895-Abstract.html}
}

@misc{Dua:2019,
  author = {Dua, Dheeru and Graff, Casey},
  institution = {University of California, Irvine, School of Information and Computer Sciences},
  title = {{UCI} Machine Learning Repository},
  url = {http://archive.ics.uci.edu/ml},
  year = 2017
}

@misc{creditcardfraud_kaggle_2013,
  author       = {Kaggle and Machine Learning Group - ULB},
  title        = {Credit Card Fraud Detection Dataset},
  year         = {2013},
  publisher    = {Kaggle},
  doi          = {10.5281/zenodo.7395559},
  url          = {https://www.kaggle.com/datasets/mlg-ulb/creditcardfraud},
  note         = {Dataset containing anonymized credit card transactions with fraud labels},
}

@article{scholkopf2001estimating,
  title={Estimating the Support of a High-Dimensional Distribution},
  author={Sch{\"o}lkopf, Bernhard and Platt, John C. and Shawe-Taylor, John and Smola, Alex J. and Williamson, Robert C.},
  journal={Neural Computation},
  volume={13},
  number={7},
  pages={1443--1471},
  year={2001},
  url={https://ieeexplore.ieee.org/abstract/document/6790022}
}

@inproceedings{liu2008isolation,
  title={Isolation Forest},
  author={Liu, Fei Tony and Ting, Kai Ming and Zhou, Zhi-Hua},
  booktitle={Proceedings of the 2008 Eighth IEEE International Conference on Data Mining},
  pages={413--422},
  year={2008},
  url={https://ieeexplore.ieee.org/abstract/document/4781136}
}

@article{liu2012isolation,
  title={Isolation-Based Anomaly Detection},
  author={Liu, Fei Tony and Ting, Kai Ming and Zhou, Zhi-Hua},
  journal={ACM Transactions on Knowledge Discovery from Data},
  volume={6},
  number={1},
  pages={3:1--3:39},
  year={2012},
  url={https://dl.acm.org/doi/abs/10.1145/2133360.2133363}
}

@inproceedings{bergman2020classification,
  title={Classification-Based Anomaly Detection for General Data},
  author={Bergman, Liron and Hoshen, Yedid},
  booktitle={International Conference on Learning Representations},
  year={2020},
  url={https://openreview.net/forum?id=H1lK_lBtvS}
}

@article{schollwock2011dmrg,
  title={The Density-Matrix Renormalization Group in the Age of Matrix Product States},
  author={Schollw{\"o}ck, Ulrich},
  journal={Annals of Physics},
  volume={326},
  number={1},
  pages={96--192},
  year={2011},
  url={https://www.sciencedirect.com/science/article/abs/pii/S0003491610001752}
}

@article{chandola2009anomaly,
  title={Anomaly Detection: A Survey},
  author={Chandola, Varun and Banerjee, Arindam and Kumar, Vipin},
  journal={ACM Computing Surveys},
  volume={41},
  number={3},
  pages={15},
  year={2009},
  url={https://dl.acm.org/doi/abs/10.1145/1541880.1541882}
}

@article{white1992dmrg,
  title={Density Matrix Formulation for Quantum Renormalization Groups},
  author={White, Steven R.},
  journal={Physical Review Letters},
  volume={69},
  number={19},
  pages={2863--2866},
  year={1992},
  publisher={American Physical Society},
  doi={10.1103/PhysRevLett.69.2863}
}

@inproceedings{xu2015appearance,
  title={Learning Deep Representations of Appearance and Motion for Anomalous Event Detection},
  author={Xu, Dan and Ricci, Elisa and Yan, Yan and Song, Jingkuan and Sebe, Nicu},
  booktitle={British Machine Vision Conference (BMVC)},
  year={2015},
  url={https://www.bmva-archive.org.uk/bmvc/2015/papers/paper008/abstract008.pdf}
}

@article{andrews2016autoencoder,
  title={Detecting Anomalous Data Using Auto-Encoders},
  author={Andrews, Jerone and Morton, Edward and Griffin, Lewis},
  journal={International Journal of Machine Learning and Computing},
  volume={6},
  number={1},
  pages={21--27},
  year={2016},
  url={https://www.ijml.org/index.php?m=content&c=index&a=show&catid=62&id=643}
}

@inproceedings{akcay2019ganomaly,
  title={Ganomaly: Semi-Supervised Anomaly Detection via Adversarial Training},
  author={Ak{\c{c}}ay, Samet and Atapour-Abarghouei, Amir and Breckon, Toby P.},
  booktitle={Asian Conference on Computer Vision (ACCV)},
  pages={622--637},
  publisher={Springer},
  year={2019},
  url={https://link.springer.com/chapter/10.1007/978-3-030-20893-6_39}
}

@inproceedings{deecke2018gan,
  title={Image Anomaly Detection with Generative Adversarial Networks},
  author={Deecke, Lucas and Vandermeulen, Robert and Ruff, Lukas and Mandt, Stephan and Kloft, Marius},
  booktitle={European Conference on Machine Learning and Knowledge Discovery in Databases (ECML-PKDD)},
  pages={3--17},
  year={2018},
  url={https://link.springer.com/chapter/10.1007/978-3-030-10925-7_1}
}

@article{donahue2016adversarial,
  title={Adversarial Feature Learning},
  author={Donahue, Jeff and Kr{\"a}henb{\"u}hl, Philipp and Darrell, Trevor},
  journal={arXiv preprint arXiv:1605.09782},
  year={2016},
  url={https://arxiv.org/abs/1605.09782}
}

@article{erfani2016high,
  title={High-Dimensional and Large-Scale Anomaly Detection Using a Linear One-Class SVM with Deep Learning},
  author={Erfani, Sarah M. and Rajasegarar, Sutharshan and Karunasekera, Shanika and Leckie, Christopher},
  journal={Pattern Recognition},
  volume={58},
  pages={121--134},
  year={2016},
  url={https://www.sciencedirect.com/science/article/abs/pii/S0031320316300267}
}

@inproceedings{goodfellow2014gan,
  title={Generative Adversarial Nets},
  author={Goodfellow, Ian J. and Pouget-Abadie, Jean and Mirza, Mehdi and Xu, Bing and Warde-Farley, David and Ozair, Sherjil and Courville, Aaron and Bengio, Yoshua},
  booktitle={Advances in Neural Information Processing Systems (NeurIPS)},
  volume={27},
  pages={2672--2680},
  year={2014},
  url={https://proceedings.neurips.cc/paper/2014/hash/f033ed80deb0234979a61f95710dbe25-Abstract.html}
}

@article{schlegl2017unsupervised,
  title={f-AnoGAN: Fast unsupervised anomaly detection with generative adversarial networks},
  author={Schlegl, Thomas and Seeb{\"o}ck, Philipp and Waldstein, Sebastian M and Langs, Georg and Schmidt-Erfurth, Ursula},
  journal={Medical image analysis},
  volume={54},
  pages={30--44},
  year={2019},
  publisher={Elsevier},
  url={https://www.sciencedirect.com/science/article/abs/pii/S1361841518302640}
}

@article{zenati2018efficient,
  title={Efficient GAN-Based Anomaly Detection},
  author={Zenati, Houssam and Foo, Chuan Sheng and Lecouat, Bruno and Manek, Gaurav and Chandrasekhar, Vijay Ramaseshan},
  journal={arXiv preprint arXiv:1802.06222},
  year={2018},
  url={https://arxiv.org/abs/1802.06222}
}

@article{schollwock2005dmrg,
  title={The Density-Matrix Renormalization Group},
  author={Schollw{\"o}ck, Ulrich},
  journal={Reviews of Modern Physics},
  volume={77},
  number={1},
  pages={259--315},
  year={2005},
  publisher={American Physical Society},
  doi={10.1103/RevModPhys.77.259}
}

@article{orus2014tensor,
  title={A Practical Introduction to Tensor Networks: Matrix Product States and Projected Entangled Pair States},
  author={Or{\'u}s, Rom{\'a}n},
  journal={Annals of Physics},
  volume={349},
  pages={117--158},
  year={2014},
  publisher={Elsevier},
  doi={10.1016/j.aop.2014.06.013}
}

@article{seebock2019unsupervised,
  title={Unsupervised Identification of Disease Marker Candidates in Retinal OCT Imaging Data},
  author={Seeb{\"o}ck, Philipp and Waldstein, Sebastian M. and Klimscha, Sophie and Bogunovi{\'c}, Hrvoje and Schlegl, Thomas and Gerendas, Bianca S. and Donner, Rene and Schmidt-Erfurth, Ursula and Langs, Georg},
  journal={IEEE Transactions on Medical Imaging},
  volume={38},
  number={4},
  pages={1037--1047},
  year={2019},
  publisher={IEEE},
  doi={10.1109/TMI.2018.2871372}
}

@article{ZUNKOVIC2023126556,
title = {Positive unlabeled learning with tensor networks},
journal = {Neurocomputing},
volume = {552},
pages = {126556},
year = {2023},
issn = {0925-2312},
doi = {https://doi.org/10.1016/j.neucom.2023.126556},
url = {https://www.sciencedirect.com/science/article/pii/S0925231223006793},
author = {Bojan Žunkovič}
}

@article{Guo2018MPOlearning,
  title = {Matrix product operators for sequence-to-sequence learning},
  author = {Guo, Chu and Jie, Zhanming and Lu, Wei and Poletti, Dario},
  journal = {Phys. Rev. E},
  volume = {98},
  issue = {4},
  pages = {042114},
  numpages = {12},
  year = {2018},
  month = {Oct},
  publisher = {American Physical Society},
  doi = {10.1103/PhysRevE.98.042114},
  url = {https://link.aps.org/doi/10.1103/PhysRevE.98.042114}
}

@article{Casagrande2024PCA,
  title = {Tensor-networks-based learning of probabilistic cellular automata dynamics},
  author = {Casagrande, Heitor P. and Xing, Bo and Munro, William J. and Guo, Chu and Poletti, Dario},
  journal = {Phys. Rev. Res.},
  volume = {6},
  issue = {4},
  pages = {043202},
  numpages = {8},
  year = {2024},
  month = {Nov},
  publisher = {American Physical Society},
  doi = {10.1103/PhysRevResearch.6.043202},
  url = {https://link.aps.org/doi/10.1103/PhysRevResearch.6.043202}
}

@article{Guo2020NonMarkovian,
  title = {Tensor-network-based machine learning of non-Markovian quantum processes},
  author = {Guo, Chu and Modi, Kavan and Poletti, Dario},
  journal = {Phys. Rev. A},
  volume = {102},
  issue = {6},
  pages = {062414},
  numpages = {8},
  year = {2020},
  month = {Dec},
  publisher = {American Physical Society},
  doi = {10.1103/PhysRevA.102.062414},
  url = {https://link.aps.org/doi/10.1103/PhysRevA.102.062414}
}

@article{HanZhang2018,
  title = {Unsupervised Generative Modeling Using Matrix Product States},
  author = {Han, Zhao-Yu and Wang, Jun and Fan, Heng and Wang, Lei and Zhang, Pan},
  journal = {Phys. Rev. X},
  volume = {8},
  issue = {3},
  pages = {031012},
  numpages = {13},
  year = {2018},
  month = {Jul},
  publisher = {American Physical Society},
  doi = {10.1103/PhysRevX.8.031012},
  url = {https://link.aps.org/doi/10.1103/PhysRevX.8.031012}
}

@article{oseledets2011tt,
  title={Tensor-Train Decomposition},
  author={Oseledets, Ivan V.},
  journal={SIAM Journal on Scientific Computing},
  volume={33},
  number={5},
  pages={2295--2317},
  year={2011},
  publisher={SIAM},
  doi={10.1137/090752286}
}

@article{novikov2017exponentialmachines,
      title={Exponential Machines}, 
      author={Alexander Novikov and Mikhail Trofimov and Ivan Oseledets},
      year={2017},
      journal={arXiv preprint arXiv:1605.03795},
      url={https://arxiv.org/abs/1605.03795}, 
}

@article{cichocki2014erabigdataprocessing,
  title={Era of big data processing: A new approach via tensor networks and tensor decompositions},
  author={Cichocki, Andrzej},
  journal={arXiv preprint arXiv:1403.2048},
  year={2014},
  url={https://arxiv.org/abs/1403.2048}, 
}

@article{SornsaengPoletti2026,
  title = {Exploring the performance of superposition of product states: From one-dimensional to three-dimensional quantum spin systems},
  author = {Sornsaeng, Apimuk and Arad, Itai and Poletti, Dario},
  journal = {Phys. Rev. E},
  volume = {113},
  issue = {3},
  pages = {035309},
  numpages = {11},
  year = {2026},
  month = {Mar},
  publisher = {American Physical Society},
  doi = {10.1103/2p2k-3j2s},
  url = {https://link.aps.org/doi/10.1103/2p2k-3j2s}
}

\appendix

\section{Physical validation}\label{app: phase}

To validate the effect of uncorrelated features in the training on SMT-AD, we train the SMT-AD model to learn the spin configurations drawn from two different ground states of the transverse-field Ising model in different phases. These two ground states are the ground states in a near-critical phase ($h_x=2.0$), which is a highly correlated state, and in a near-paramagnetic phase ($h_x=8.0$), which is an uncorrelated state, so the spin configuration drawn from these states will have correlations determined by the state probability distribution. Spin configurations are drawn systematically from the distributions by using the Metropolis-Hasting algorithm. In the training scheme, we have two different scenarios: training with near-critical samples and detecting near-paramagnetic samples, and \textit{vice versa}. For each scheme, half of the assigned normal samples will be used to train the model, while the other half will be used in the test, and a number of anomalies will be 5\% of the number of normal test samples. Note that since the spin configuration samples are just 0 and 1, the preprocessing scheme can be done directly without using rank normalization encoding. The hyperparameter scanning and the optimization schemes are done similarly as detailed in Sec. \ref{section: implement}.

Table~\ref{table: phase} shows the AUROC and AUPRC for detecting spin configurations in paramagnetic and near-critical phases as anomaly sets. In detecting paramagnetic spin configurations as anomalies, the models learn the near-critical spin configurations, which are highly correlated data. TNAD needs to use a large bond dimension to understand the highly correlated dataset and successfully detect the anomalies, while SMT-AD needs a large number of bond-dimension-one MPOs to understand this dataset, corresponding to increasing the number of bond-dimension-one MPSs to explain the near-critical ground state \cite{SornsaengPoletti2026}. This validates that the correlation structure of the training data governs the expressive power of SMT-AD.

\begin{table}[]
\caption{Average ($\pm$ standard deviation) AUROC and AUPRC in a task: detecting paramagnetic from near-critical spin configurations, from TNAD and SMT-AD with specific hyperparameters. These results are averaged over 20 realizations.}
\label{table: phase}
\begin{tabular}{|cc|cc|cc|}
\hline
 \multicolumn{2}{|c|}{Hyperparameters} & \multicolumn{2}{c|}{AUROC}         & \multicolumn{2}{c|}{AUPRC}         \\ \hline
         \multicolumn{1}{|c|}{TNAD $(\chi, P)$} & SMT-AD $(M,P)$  & \multicolumn{1}{c|}{TNAD} & SMT-AD & \multicolumn{1}{c|}{TNAD} & SMT-AD \\ \hline
\multicolumn{1}{|c|}{$(10,2)$} & $(36,3)$ & \multicolumn{1}{c|}{97.8 $\pm$ 0.2} & 79.4 $\pm$ 0.1 & \multicolumn{1}{c|}{98.6 $\pm$ 0.1} & 22.3 $\pm$ 0.1 \\\hline
\end{tabular}
\end{table}

\end{document}